\documentclass{article}
\usepackage{ijcai25}

\usepackage{times}
\usepackage{soul}
\usepackage{url}
\usepackage[hidelinks]{hyperref}
\usepackage[utf8]{inputenc}
\usepackage[small]{caption}
\usepackage{graphicx}
\usepackage{amsmath}
\usepackage{amsthm}
\usepackage{booktabs}
\usepackage{algorithm}
\usepackage{algorithmic}
\usepackage[switch]{lineno}

\title{A Survey on Image Quality Assessment:\\ Insights, Analysis, and Future Outlook}

\author{
Chengqian Ma$^1$
\and
Zhengyi Shi$^2$\and
Zhiqiang Lu$^2$\and
Shenghao Xie$^1$\and
Fei Chao$^2$\And
Yao Sui$^1$\\
\affiliations
$^1$National Institute of Health Data Science, Peking University, Beijing, China\\
$^2$School of Informatics, Xiamen University, Xiamen, Fujian, China\\
}

\begin{document}

\maketitle

\begin{abstract}
Image quality assessment (IQA) represents a pivotal challenge in image-focused technologies, significantly influencing the advancement trajectory of image processing and computer vision. Recently, IQA has witnessed a notable surge in innovative research efforts, driven by the emergence of novel architectural paradigms and sophisticated computational techniques. This survey delivers an extensive analysis of contemporary IQA methodologies, organized according to their application scenarios, serving as a beneficial reference for both beginners and experienced researchers. We analyze the advantages and limitations of current approaches and suggest potential future research pathways. The survey encompasses both general and specific IQA methodologies, including conventional statistical measures, machine learning techniques, and cutting-edge deep learning models such as convolutional neural networks (CNNs) and Transformer models. The analysis within this survey highlights the necessity for distortion-specific IQA methods tailored to various application scenarios, emphasizing the significance of practicality, interpretability, and ease of implementation in future developments.
\end{abstract}

\section{Introduction}
Methods for image quality assessment (IQA) play a critical role in establishing benchmarks to refine algorithms within various domains of computer vision. IQA is crucial to evaluate the efficacy of image processing algorithms, such as dehazing algorithms (DHAs)~\cite{min2019quality}, medical image processing methods~\cite{rosen2018quantitative}, and image deblurring algorithms~\cite{liu2013no}. These algorithms aim to improve image quality, and their success is often demonstrated through the improved quality of the images produced. Previous surveys have organized IQA methods by the availability of a reference image~\cite{5412098}, types of distortions~\cite{Xu2017NoreferenceBlindIQ}, and utilized techniques~\cite{8822415}. However, these surveys fall short in covering the wide range of application scenarios and lag behind the latest innovations. Our study provides an updated review of recent IQA advancements, focusing on different application scenarios to offer a comprehensive overview of the field.

\begin{figure}[t]
    \centering
    \includegraphics[width=0.45\textwidth]{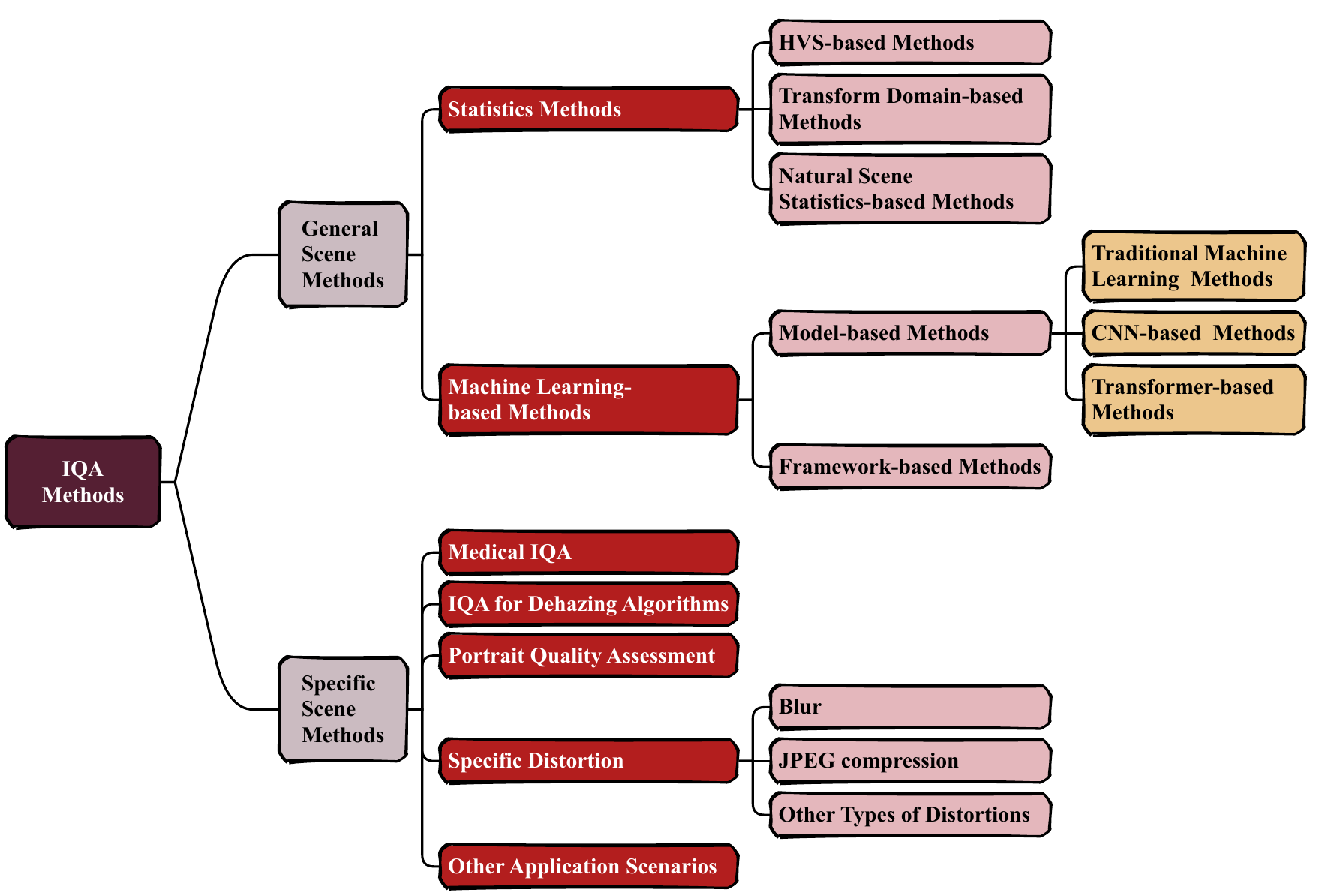}
    \caption{Classification of IQA Methods.}
    \label{fig:metric_struct}
\end{figure}

IQA can be classified into two categories: subjective IQA (SIQA) and objective IQA (OIQA). SIQA relies on human evaluators and is further subdivided by the presence or absence of a reference image. In single stimulus rating~\cite{series2012methodology}, evaluators assign scores based solely on their personal judgment, whereas in double stimulus~\cite{series2012methodology} rating, a reference image is provided for comparison. OIQA is an automated process that excludes human intervention and is typically evaluated using distorted images with corresponding mean opinion scores (MOSs) determined by human evaluation.
Although SIQA is the benchmark due to its high alignment with the human visual system (HVS), it is often deemed impractical for extensive application as it is time-consuming, expensive, and highly subjective. In contrast, OIQA methods are fast and cost-effective. The main challenge lies in developing OIQA techniques to achieve accuracy and reliability comparable to SIQA. OIQA methods are further classified into full-reference (FR), reduced-reference (RR), and no-reference (NR) categories. The NR category, also known as blind image quality assessment (BIQA), has attracted considerable research focus due to the absence of need for reference images in various applications.

Different application scenarios impose distinct requirements on IQA methods. For example, medical imaging prioritizes lesion visibility, while portrait photography emphasizes aesthetics. These varying, and occasionally conflicting, demands call for specialized IQA approaches that are context-specific. This paper initially examines general IQA methods that are applicable in various scenarios before diving into those tailored for particular applications. Table~\ref{tab:abbreviations} lists the specialized terms and their abbreviations used in this paper, while Figure~\ref{fig:metric_struct} depicts the categorization of the IQA methods presented in this survey.


\begin{table}[t]
    \centering
    \begin{tabular}{ll}
        \toprule
        \textbf{Abbreviations} & \textbf{Full Name} \\
        \midrule
        IQA & Image Quality Assessment \\
        SVD & Singular Value Decomposition \\
        HVS & Human Visual System \\
        SIQA & Subjective IQA \\
        OIQA & Objective IQA \\
        BIQA & Blind Image Quality Assessment \\
        NAR & Non-Aligned Reference \\
        AR & Aligned Reference \\
        MOS & Mean Opinion Score \\
        CNN & Convolutional Neural Network \\
        GAN & Generative Adversarial Networks \\
        PC & Phase Congruency \\
        GM & Gradient Magnitude \\
        DCT & Discrete Cosine Transform \\
        NSS & Natural Scene Statistics \\
        HDR & High Dynamic Range \\
        LDR & Low Dynamic Range \\
        TMOs & Tone-Mapped Images Operators \\
        DHA & Dehazing Algorithm \\
        CNR & Contrast-to-Noise Ratio \\
        SNR & Signal-to-Noise Ratio \\
        ViT & Vision Transformer \\
        \bottomrule
    \end{tabular}
    \caption{Terminology and Abbreviation}
    \label{tab:abbreviations}
\end{table}

\section{Image Quality Assessment Methods}
\subsection{General Scene Methods}
This section outlines the IQA methods applicable to general scenes, divided into two segments: \textit{Statistics Methods} and \textit{Machine Learning-based Methods}.
\subsubsection{Statistics Methods}
Statistics methods predominantly encompass \textit{HVS-based Methods}, \textit{Transform Domain-based Methods}, and \textit{Natural Scene Statistics (NSS)-based Methods}.
\paragraph{HVS-Based Methods}
The fundamental metrics, such as mean squared error (MSE), signal-to-noise ratio (SNR), peak signal-to-noise ratio (PSNR), and universal image quality index (UQI)~\cite{wang2002universal}, evaluate image noise and fidelity based on statistical characteristics and overlook the HVS. This oversight leads to notable discrepancies from human perception. In response, methodologies like Visual SNR (VSNR)~\cite{chandler2007vsnr} and HVS-based PSNR (PSNR-HVS)~\cite{egiazarian2006new} have been developed to incorporate HVS considerations. The Visual Saliency-induced Index (VSI)~\cite{zhang2014vsi} posits that the saliency of each pixel affects its importance, determined by the content of the image. This implies prioritizing the information underlying the image rather than treating every pixel equally. For example, Visual Information Fidelity (VIF)~\cite{sheikh2006image} correlates image quality with the fidelity of image information, using an information-theoretic framework to measure how distortion affects image quality.

In the course of aligning with the HVS, multiple characteristics have been identified and utilized in the design of IQA:

Gradient Magnitude Similarity Deviation (GMSD)~\cite{xue2013gradient} notes that HVS is highly sensitive to edge details, which can be captured by gradient magnitude (GM) for quality assessment. This utilization is also seen in studies such as the perceptual similarity (PSIM)~\cite{gu2017fast}, which measures the similarities of micro- and macro-structures described by GM maps; the gradient similarity (GSIM)~\cite{liu2011image} index, which introduces a measure for gradient similarity; and the feature similarity (FSIM)~\cite{zhang2011fsim} index, which selects phase congruency and gradient deviation (GD) as predictive features of quality.

Structural Similarity (SSIM)~\cite{wang2004image} focuses on the structural information of images, inspired by the HVS's key role in capturing structure within the visual scene~\cite{wang2002image}. To enhance its applicability and efficacy, variants such as multiscale SSIM (MS-SSIM)~\cite{wang2003multiscale} and information content weighted SSIM (IW-SSIM)~\cite{wang2010information} have been proposed.

The HVS employs distinct strategies when processing images of varying qualities. For high-quality images, the HVS focuses more on point-wise variations, rendering metrics like MSE and PSNR appropriate. In the case of lower-quality images, the HVS prioritizes whether the image structure and content remain intact. The Most Apparent Distortion (MAD)~\cite{larson2010most} method addresses this by considering both strategies and weighting their scores according to image quality. Table~\ref{tab:hvs_based} summarizes the HVS-based methods.

\begin{table}
    \centering
    \begin{tabular}{lrr}
        \toprule
        Metric & Citation & Time \\
        \midrule
        SSIM   & 58,352 & 2004.04 \\
        MS-SSIM & 8,043 & 2004.05 \\
        UQI    & 7,419 & 2002.03 \\
        FSIM   & 5,512 & 2011.01 \\
        VIF    & 4,861 & 2006.02 \\
        MAD    & 2,250 & 2010.01 \\
        GMSD   & 1,716 & 2013.12 \\
        VSNR   & 1,578 & 2007.08 \\
        IW-SSIM & 1,533 & 2010.11 \\
        VSI    & 1,082 & 2014.08 \\
        GSIM   & 872  & 2011.11 \\
        PSNR-HVS & 433 & 2006.01 \\
        PSIM   & 238 & 2017.05 \\
        \bottomrule
    \end{tabular}
    \caption{HVS-based Methods. The citation number is sourced from Google Scholar as of Feb. 1, 2025. The time represents the earliest appearance of the metric.}
    \label{tab:hvs_based}
\end{table}

\paragraph{Transform Domain-Based Methods}
In addition to leveraging features like structural information, edge details, and phase congruency pertinent to the HVS for IQA, images can be transformed into alternative domains where their transformed coefficients serve to characterize the image features.

Singular Value Decomposition (SVD)~\cite{shnayderman2006svd} is used to extract features for IQA at both global and local scales. In addition, methods such as SFF~\cite{chang2013sparse} and QASD~\cite{li2016sparse} employ sparse representation to derive features, while the wavelet transform is utilized in studies such as~\cite{wang2005reduced} to assess images based on changes in wavelet coefficients due to distortion. The Blind Image Integrity Notator using DCT statistics (BLIINDS-II)~\cite{saad2012blind} leverages the statistical properties of coefficients in the discrete cosine transform (DCT) domain, which are affected by distortion, for IQA purposes. Table~\ref{tab:transform_domain} provides an overview of transform domain-based methods.

\begin{table}
    \centering
    \begin{tabular}{lrr}
        \toprule
        Metric & Citation & Time \\
        \midrule
        BLIINDS-II & 1,912 & 2012.03 \\
       ~\cite{wang2005reduced}         & 600  & 2005.03 \\
       ~\cite{shnayderman2006svd}         & 482  & 2006.02 \\
        SFF        & 180  & 2013.06 \\
        QASD       & 50   & 2016.06 \\
        \bottomrule
    \end{tabular}
    \caption{Transform Domain-based Methods. The citation number is sourced from Google Scholar Feb. 1, 2025. The time represents the earliest appearance of the metric.}
    \label{tab:transform_domain}
\end{table}

\begin{table}[b]
    \centering
    \begin{tabular}{lrr}
        \toprule
        Metric & Citation & Time \\
        \midrule
        IFC    & 1,721 & 2005.11 \\
        TMQI   & 720  & 2012.01 \\
       ~\cite{gabarda2007blind}     & 397  & 2007.09 \\
        DRIM   & 371  & 2008.08 \\
        \bottomrule
    \end{tabular}
    \caption{NSS-based Methods. The citation number is sourced from Google Scholar Feb. 1, 2025. The time represents the earliest appearance of the metric.}
    \label{tab:nss_based}
\end{table}

\paragraph{NSS-Based Methods}
Alongside HVS and transform-domain features, intrinsic characteristics of natural images are useful for IQA. Methods such as IFC~\cite{sheikh2005information} analyze the statistical properties of natural images to identify distortions and assess their quality. The anisotropy-based method~\cite{gabarda2007blind} evaluates image quality by examining the variance of the expected entropy in the image, showing high consistency with human visual preferences.

The range of luminance in natural images is another significant area of interest. High dynamic range (HDR) images capture a wide range of luminance levels, but standard displays generally possess a limited dynamic range. To render HDR images on standard devices, Tone-Mapped Image Operators (TMOs) are used to convert HDR images to low dynamic range (LDR) images. Different TMOs produce varying results, and traditional FR-IQA methods are typically limited to image pairs with similar dynamic ranges. The DRIM~\cite{aydin2008dynamic} technique addresses this issue by introducing the High Dynamic Range Visible Differences Predictor (HDR-VDP) model to predict the visibility of contrast differences, effectively handling image pairs with significant dynamic range disparities and broadening the applications of IQA. The Tone-Mapped Image Quality Index (TMQI)~\cite{yeganeh2012objective}, based on SSIM~\cite{wang2004image} and naturalness metrics, achieves a comparable effect. Table~\ref{tab:nss_based} summarizes the NSS-based methods.

\subsubsection{Machine Learning-Based Methods}
This section presents Machine Learning-based Methods, focusing on two perspectives: \textit{Model-based Methods} and \textit{Framework-based Methods}.
\paragraph{Model-based Methods}
We organize Model-based Methods into three categories: \textit{Traditional Machine Learning}, \textit{CNN-based} and \textit{Transformer-based Methods}.
\subparagraph{Traditional Machine Learning Methods}
The advancement of machine learning has led to the implementation of numerous approaches in IQA. For example, methods such as MMF~\cite{liu2012image} and ParaBoost~\cite{liu2015paraboost} integrate multiple quality assessment methods tailored for distinct types of distortion, using SVR to combine scores from multiple quality metrics.

Various machine learning methods employing NSS have been proposed. For example, DIIVINE~\cite{moorthy2011blind} incorporates NSS feature extraction alongside SVM and SVR to identify distortions and evaluate quality.  IL-NIQE~\cite{zhang2015feature} constructs a multivariate Gaussian (MVG) model using NSS features for IQA. BRISQUE~\cite{mittal2012no} achieves multi-scale NSS feature extraction, distortion type analysis, and quality assessment while maintaining minimal computational complexity. Table~\ref{tab:traditional_ml} encapsulates these Traditional Machine Learning Methods.
\begin{table}
    \centering
    \begin{tabular}{lrr}
        \toprule
        Metric & Citation & Time \\
        \midrule
        BRISQUE & 5,733 & 2012.08 \\
        DIIVINE & 2,003 & 2011.01 \\
        IL-NIQE & 1,214 & 2015.04 \\
        MMF    & 216  & 2012.12 \\
        SVDR   & 184  & 2011.09 \\
        ParaBoost & 86 & 2015.12 \\
        \bottomrule
    \end{tabular}
    \caption{Traditional Machine Learning Methods. The citation number is sourced from Google Scholar Feb. 1, 2025. The time represents the earliest appearance of the metric.}
    \label{tab:traditional_ml}
\end{table}

\subparagraph{CNN-Based Methods}
Since 2014, when IQA-CNN~\cite{Kang_2014_CVPR} was first introduced, there has been a growing interest in the use of CNNs for IQA. Various CNN techniques have been successfully adapted for IQA, yielding impressive results.~\cite{zhang2018unreasonable} provides experimental evidence showing that deep features outperform alternative metrics. Some notable methods include:

BIECON~\cite{kim2016fully} divides an image into patches and merges their features to compute the quality score, serving as inspiration for subsequent studies.~\cite{bosse2017deep} proposes a completely data-driven method using a Siamese network. MEON~\cite{ma2017end} implements an end-to-end model from image to quality assessment. RankIQA~\cite{liu2017rankiqa} uses a Siamese network to learn the relative ranking of images of different qualities.

In the HVS, the top-down perception model is important for visual tasks, indicating that image content should be understood before IQA. This is because different image contents lead to different evaluation criteria, and the importance of global features can only be determined post image content comprehension. Therefore, a BIQA approach using a Self-Adaptive Hyper Network~\cite{su2020blindly} captures image semantics and constructs a perception rule using a hyper network for IQA.

IQA algorithms must incorporate global and local distortions.~\cite{varga2020multi} leverages the Inception module, which comprises convolutional kernels of various sizes, to interpret visual data at different scales simultaneously. This parallel multi-scale processing is effective since IQA algorithms must address both global structures and local texture details of an image. Since 2020, images generated by Generative Adversarial Networks (GAN) have often exhibited sharper edges and texture-like noises, which are difficult to measure accurately with traditional metrics. Parallel analysis of global and local features can address this issue, with large scales handling global structures and small scales catering to local details. From this perspective, the IQMA Network~\cite{guo2021iqma}, the winner of the NTIRE 21 IQA public leaderboard, proposes a bilateral-branch multi-scale IQA network. The network has two branches that extract multi-scale features from the reference and distorted images, respectively, and features of matching scales from these branches are fed to multiple scale-specific feature fusion modules. Re-IQA~\cite{saha2023re} also tackles this using a dual-encoder Mixture of Experts structure to capture global content alongside local quality of images. In particular, the encoding of global content uses contrastive learning in a creative way.

Most CNN-based FR IQA methods exhibit excessive sensitivity towards texture similarity. DISTS~\cite{ding2020image}is pioneering in its integration of a built-in mechanism for handling texture resampling, effectively combining structural and textural information by leveraging VGG for texture extraction. Table~\ref{tab:cnn_based} summarizes CNN-based methods.
\begin{table}
    \centering
    \begin{tabular}{lrr}
        \toprule
        Metric & Citation & Time \\
        \midrule
        IQA-CNN & 1,406 & 2014.09 \\
       ~\cite{bosse2017deep}     & 1,229 & 2017.10 \\
        DISTS  & 847  & 2020.12 \\
       ~\cite{su2020blindly} & 667 & 2020.06 \\
        MEON   & 593  & 2017.11 \\
        RankIQA & 556  & 2017.07 \\
        BIECON & 487  & 2016.12 \\
        Re-IQA & 84   & 2023.04 \\
       ~\cite{varga2020multi} & 44 & 2020.02 \\
        IQMA Network & 27 & 2021.06 \\
        \bottomrule
    \end{tabular}
    \caption{CNN-based Methods. The citation number is sourced from Google Scholar Feb. 1, 2025. The time represents the earliest appearance of the metric.}
    \label{tab:cnn_based}
\end{table}

\begin{table}[ht]
    \centering
    \begin{tabular}{lrr}
        \toprule
        Metric & Citation & Time \\
        \midrule
        MUSIQ & 601 & 2021.08 \\
       ~\cite{golestaneh2022no} & 307 & 2021.08 \\
        Maniqa & 301 & 2022.04 \\
        TRIQ & 230 & 2020.12 \\
        IQT & 167 & 2021.04 \\
        NTIRE 2021 IQA & 114 & 2021.05 \\
        NTIRE 2022 IQA & 113 & 2022.06 \\
        \bottomrule
    \end{tabular}
    \caption{Transformer-based Methods. The citation number is sourced from Google Scholar Feb. 1, 2025. The time represents the earliest appearance of the metric.}
    \label{tab:transformer_based}
\end{table}

\subparagraph{Transformer-Based Methods}
The development of Transformers has heralded a new era in Artificial Intelligence, significantly invigorating IQA. While CNNs necessitate fixed input dimensions for IQA, which requires scaling different-resolution images and leads to data loss, TRIQ~\cite{you2021transformer} uses Transformers to efficiently manage images of different resolutions. The multi-head attention mechanism in the Transformer framework enhances the focus on global features, surpassing the capabilities of CNNs. Building on this, MUSIQ~\cite{ke2021musiq} introduces an innovative method using a hash-based 2D spatial embedding along with a scale embedding. This approach facilitates the processing of images with varying sizes and aspect ratios without necessitating cropping or resizing, which is crucial for IQA applications involving resolution- and aspect-ratio-sensitive images.

Various Vision Transformer (ViT)-based methods have achieved impressive results. For example, the IQT~\cite{cheon2021perceptual} method uses a Siamese architecture to extract features from reference and distorted images, using ViT to evaluate their quality. ViT with relative ranking and self-consistency~\cite{golestaneh2022no} introduces two new losses during the ViT process. The first loss emphasizes the importance of relative image quality scores over absolute scores to better exploit relative ranking information. The second loss addresses the issue of diminished performance in NR-IQA models when images undergo equivariant transformations (e.g., horizontal flipping). Maniqa~\cite{yang2022maniqa} extracts features using ViT and feeds them into the Transposed Attention Block and the Scale Swin Transformer Block, which employ attention mechanisms to analyze features across channels and spatial dimensions.

In the NTIRE 2021 Challenge on Perceptual IQA~\cite{royat2021ntire}, several teams adopted the Transformer framework. The LIPT team pioneered its application to the FR-IQA task. In the following NTIRE 2022 Challenge on Perceptual IQA~\cite{gu2022ntire}, the winners of the FR-IQA and NR-IQA tracks used ViT frameworks, indicating a broad acceptance of ViT-based IQA methods due to their proven effectiveness. Table~\ref{tab:transformer_based} summarizes Transformer-based methods.

\paragraph{Framework-Based Methods}
Besides advancements in model architecture, numerous studies in machine learning focus on developing new training frameworks to address crucial challenges in the IQA field.

Neural networks typically require extensive datasets for effective training, yet numerous specialized IQA tasks face data scarcity. Therefore, CNN-based Medical Ultrasound IQA~\cite{zhang2021cnn} uses transfer learning to leverage the knowledge from existing datasets as a foundation to improve the performance of the Medical Ultrasound IQA model. MetaIQA~\cite{zhu2020metaiqa}, on the other hand, aggregates an extensive dataset featuring various distortion types, mirroring the meta-knowledge that humans use to assess different distortions, which is then applied to evaluate unknown distortion types. UNIQUE~\cite{zhang2021uncertainty}, a BIQA model, demonstrates cross-distortion generalization and proposes a method for simultaneous training on multiple IQA databases. This approach allows the model trained on one distortion type to use knowledge from other distortion types. DeepFL-IQA~\cite{lin2020deepfl} introduces a weakly supervised learning strategy that uses objective IQA metric scores as references during training, with subsequent fine-tuning using subjective quality scores to ensure predictions align with HVS responses, thus mitigating data scarcity issues. CONTRIQUE~\cite{madhusudana2022image} and QAC~\cite{xue2013learning} use self-supervised learning to train on unlabeled images without human scoring. Hallucinated-IQA~\cite{lin2018hallucinated} adopts adversarial learning for NR-IQA. Collectively, these methods effectively tackle the prevalent issue of data scarcity.

A critical issue in NR-IQA research is that CNNs process images pixel-by-pixel without knowing the score for each pixel, in contrast to FR-IQA where each pixel in the reference image is matched with a corresponding pixel for evaluation. Therefore, integrating the knowledge from FR-IQA into NR-IQA is crucial to provide references for each pixel. As a solution,~\cite{kim2018deep} proposes a two-stage training framework. In the initial stage, a CNN predicts an objective error map by comparing reference and distorted images, using the error map as a proxy training target. This approach increases training data and mitigates overfitting. In the second stage, the model is refined to predict human subjective scores. CVRDK-IQA~\cite{yin2022content} accomplishes this using knowledge distillation, which introduces multiple high-quality image priors through non-aligned reference (NAR) images. The disparity in distribution between high- and low-quality images allows the model to better assess image quality. Knowledge distillation transfers this disparity from the FR-teacher to the NAR-student, enabling the student to appreciate the characteristics of high-quality images in the context of IQA. Table~\ref{tab:framework_based} summarizes framework-based methods.
\begin{table}
    \centering
    \begin{tabular}{lrr}
        \toprule
        Metric & Citation & Time \\
        \midrule
        QAC & 470 & 2013.06 \\
        MetaIQA & 405 & 2020.04 \\
       ~\cite{kim2018deep} & 315 & 2018.06 \\
        UNIQUE & 286 & 2021.03 \\
        Hallucinated-IQA & 273 & 2018.04 \\
        CONTRIQUE & 206 & 2022.06 \\
        DeepFL-IQA & 67 & 2020.01 \\
        CVRDK-IQA & 34 & 2022.02 \\
        Medical Ultrasound IQA & 28 & 2021.07 \\
        \bottomrule
    \end{tabular}
    \caption{Framework-based Methods. The citation number is sourced from Google Scholar Feb. 1, 2025. The time represents the earliest appearance of the metric.}
    \label{tab:framework_based}
\end{table}

\begin{table}
    \centering
    \begin{tabular}{lrr}
        \toprule
        Metric & Citation & Time \\
        \midrule
       ~\cite{ferzli2009no} & 1,004 & 2009.04 \\
       ~\cite{rosen2018quantitative} & 360 & 2018.04 \\
        BIBLE & 280 & 2015.01 \\
       ~\cite{min2019quality} & 222 & 2019.02 \\
       ~\cite{liu2013no} & 127 & 2013.11 \\
       ~\cite{gore2015full} & 40 & 2015.02 \\
       ~\cite{chahine2024deep} & 15 & 2024.04 \\
        \bottomrule
    \end{tabular}
    \caption{Specific Scene Metrics. The citation number is sourced from Google Scholar Feb. 1, 2025. The time represents the earliest appearance of the metric.}
    \label{tab:specific_scene}
\end{table}
\subsection{Specific Scene Methods}
Besides the general IQA methods, numerous specialized methods have been developed for specific scenarios. These methods are customized to address the unique demands of their respective applications. Table~\ref{tab:specific_scene} summarizes the Specific Scene Methods.
\subsubsection{Medical IQA}
In brain imaging research, data quality is a significant factor, especially in studies focusing on brain development, where age correlates with data quality.~\cite{rosen2018quantitative} evaluated various quantitative metrics for image quality, including those from the Preprocessed Connectomes Project's Quality Assurance Protocol and the Euler number from FreeSurfer. Among these, the Euler number emerged as the most effective. This result arises because metrics like SNR and Contrast-to-Noise Ratio (CNR) assess global image characteristics, whereas the Euler number examines local topological structures. Local motion artifacts or reconstruction errors may have a minor impact on global metrics but can significantly affect the Euler number. For example, a local motion artifact can create holes or discontinuities in the cortical surface, greatly impacting the Euler number but minimally influencing the SNR or CNR. Consequently, selecting appropriate quality assessment metrics in medical applications should be tailored to specific requirements of the study and the characteristics of the data.
\subsubsection{IQA for Dehazing Algorithms}
DHAs aim to improve image quality in challenging weather conditions like fog by eliminating haze and restoring clarity, contrast, and color, thereby enhancing visual quality and usability. Traditional FR IQA methods fall short in evaluating DHAs effectively because dehazing involves not only image restoration but also contrast enhancement, color adjustment, and structural recovery. These characteristics limit the effectiveness of traditional FR-IQA methods in assessing dehazing outcomes.

~\cite{min2019quality} proposes a method for evaluating dehazing algorithms, emphasizing a holistic approach to image structure restoration, color rendition, and over-enhancement of low-contrast areas comprehensively. This method effectively adapts to the contrast enhancement and color adjustment in dehazing, significantly improving its applicability and accuracy, particularly in the context of aerial images.
\subsubsection{Portrait Quality Assessment}
With the widespread use of smartphones, portrait photography has become ubiquitous, yet assessing the quality of these images remains a challenge. In portraits, the facial region is typically the visual focus, and its quality has a greater impact on overall portrait quality than other areas. Traditional IQA methods usually perform global assessments on the entire image, which fails to adequately emphasize the importance of the facial region. Moreover, portrait quality depends not only on the facial region but also on global factors such as background and composition. Traditional IQA methods often struggle to balance local and global quality assessments simultaneously. Therefore, there is a need for specialized methods to assess portrait quality.

In the NTIRE 2024 Challenge, Deep Portrait Quality Assessment track~\cite{chahine2024deep}, each team introduced unique methods. For example, PQE~\cite{chahine2024deep} proposed a two-branch portrait quality assessment model, focusing separately on the entire image and facial components, thus capturing distinct scene and quality features. SAR~\cite{chahine2024deep} devised a network that integrates scene-adaptive global context and local face-awareness, using a face detector for precise facial localization and employing a ViT to model local facial details and the global image. These methods emphasize separate processing of facial features and the background, providing targeted IQA solutions.
\subsubsection{Specific Distortion}
In computer vision and image processing, image deblurring is a crucial task that aims to recover a sharp image from its blurry counterpart. Blurs can be caused by various factors such as camera shake, object motion, or inaccurate focus. The goal of deblurring algorithms is to eliminate these blurs and restore the original image. However, it is often difficult to perfectly recover the ideal clear image in practice. These algorithms may produce various artifacts that detract from image quality. For example, ringing artifacts (ripple-like structures near edges) are prevalent in deblurring processes and can significantly disrupt human visual perception.

Various deblurring algorithms may produce unique artifact types, each with distinct characteristics. Therefore, a method that can effectively evaluate these artifacts is needed.~\cite{liu2013no} designs a series of features specifically targeting deblurring artifacts, including a new NR method to detect large-scale ringing artifacts; various methods for gauging noise levels in deblurred results; and multiple sharpness metrics for assessing deblurring clarity. These features collectively allow for a thorough assessment of deblurring quality across different artifact types, rather than focusing solely on a single type.

Blurring also causes the attenuation of high-frequency components in images, altering the size of Tchebichef moments. Therefore, BIBLE~\cite{li2015no} calculates gradients to remove low-frequency components, allowing high-frequency components to dominate and more effectively represent blurring. It then uses Tchebichef moments computed from gradient images to assess blur.

Various IQA techniques focus on distinct factors that cause image degradation. For example,~\cite{ferzli2009no} incorporates the perceptual characteristics of HVS directly into a sharpness metric based on Just Noticeable Blur (JNB), achieving alignment of the sharpness measurement with human subjective perception. In JPEG compression, issues like blocking effect and blurring effect require significant attention.~\cite{gore2015full} evaluates the quality of JPEG compressed images, effectively capturing these distortions and outperforming conventional IQA methods.

In summary, a study of the IQA methods in various settings clearly indicates that IQA must meet specific requirements. Therefore, suggesting a universal IQA method that addresses all challenges is impractical. Instead, the focus should shift towards scene-specific IQA, considering the unique demands when designing IQA metrics.

\begin{figure*}[htbp]
    \centering
    \includegraphics[width=\textwidth]{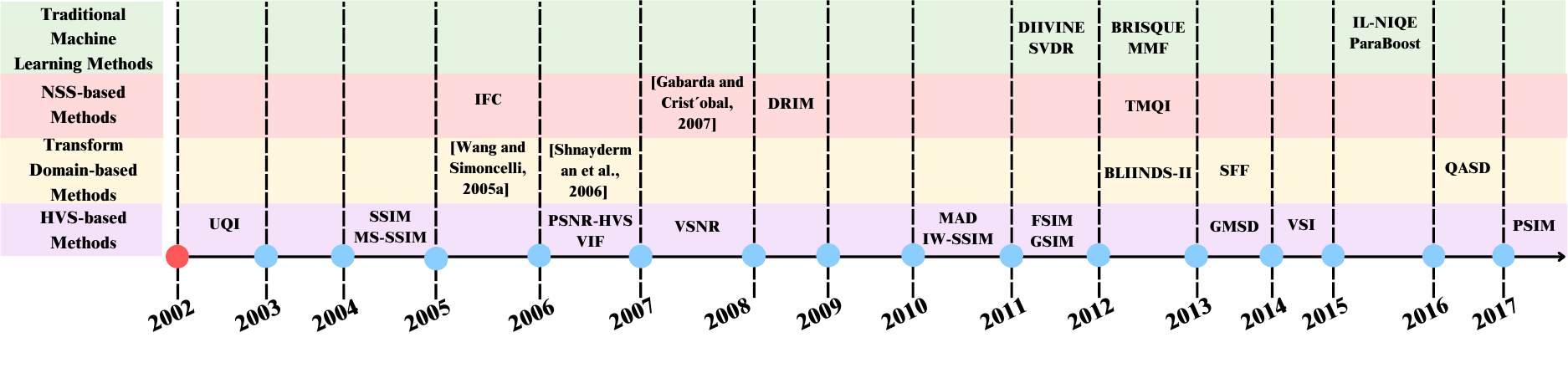}
    \caption{Publication Times of HVS-based Methods, Transform Domain-based Methods, NSS-based Methods and Traditional Machine Learning Methods. For methods that span more than one line, we enclose them in [].}
    \label{fig:early}
\end{figure*}

\begin{figure*}[htbp]
    \centering
    \includegraphics[width=\textwidth]{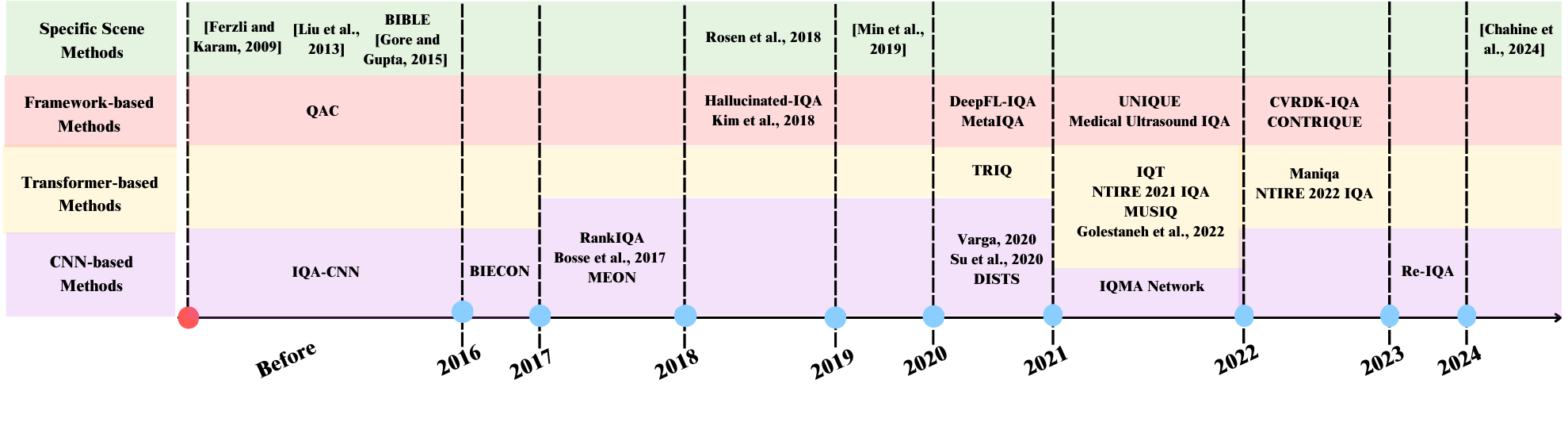}
    \caption{Publication Times of CNN-Based Methods, Transformer-Based Methods, Framework-Based Methods, Specific Scene Methods. For methods that span more than one line, we enclose them in [].}
    \label{fig:later}
\end{figure*}

\section{Discussion and Analysis}
Figures~\ref{fig:early} and~\ref{fig:later} provide a chronological overview of the IQA methods discussed in this paper. Over the course of IQA advancement, primary technologies have progressed from basic statistical indices to traditional machine learning methods like SVR-based models, then to deep learning methods such as CNNs, Transformer-based models, and various training frameworks like meta-learning. However, the predominant IQA methods in use continue to be the traditional PSNR and SSIM, primarily due to their straightforward nature  and high interpretability.

In terms of simplicity, using the top-performing ViT-based model for IQA requires researchers to navigate numerous technical challenges in environment setup and model deployment. Regarding interpretability, neural network-based methods function as black boxes, potentially leading to situations where favorable IQA by a ViT-based module in an image generation system might stem from the image aligning with the IQA's scoring preferences rather than the intrinsic quality of the image.

In the development of IQA, it is imperative to recognize that IQA technology must address diverse practical application scenarios and align with user requirements. The evaluation criteria for image quality should be derived from the perspective of image users. For example, assessing the aesthetic value of an image should be approached from an aesthetic point of view, using standards such as composition, lighting, focus control, and color, as suggested by~\cite{luo2008photo}. When evaluating portrait quality, the approach should mirror that of the teams participating in the NTIRE 2024 Portrait Quality Assessment Challenge~\cite{chahine2024deep}, such as PQE, which evaluates the different effects of facial and background components on portrait quality, recommending a two-branch portrait quality assessment model. Similarly, the model from team SECE-SYSU introduces Scene Adaptive Regressors to separately consider different scenes in portrait quality IQA.

Ultimately, proposing a universal IQA method applicable across all scenarios is impractical, as different scenes demand contrasting quality criteria. For example, motion blur might enhance the realism of an image and be desirable in portrait photography, yet it is deemed poor quality in medical diagnostic images. Future development of IQA methods should begin by focusing on the specific application domains, considering the characteristics of images in those fields and the needs of associated algorithms. Simultaneously, it is crucial to ensure that the metrics are user-friendly and highly interpretable to facilitate their application.
\section{Conclusion}
This paper systematically reviews the main research advancements in the field of IQA, covering both general and specific scenarios, and transitioning from traditional methods to advanced deep learning-based techniques. It offers a comprehensive overview of the development of IQA technology. Through the analysis of the strengths and limitations of various methods, it is apparent that despite continuous advancements, the preference still leans towards traditional methods like PSNR and SSIM. This is mainly due to the simplicity and high interpretability of traditional methods. In contrast, although deep learning-based methods excel in performance, they encounter technical barriers and are often less interpretable in practical applications.

Future advances in IQA technology should prioritize specific scenario needs and create specialized IQA methods suited to varied application contexts. In medical imaging, for example, image quality assessment should focus on lesion visibility, whereas in portrait photography, the quality of the facial area significantly influences the overall portrait quality over other regions. Moreover, the application of IQA technology in production practice should align with user requirements, considering the criteria for assessing image quality from the perspective of image users. When assessing image quality, researchers developing dehazing algorithms ought to prioritize factors such as contrast enhancement and color adjustment, whereas those investigating deblurring algorithms need to be particularly attentive to artifacts such as ringing. 

Consequently, advancing IQA technology requires focusing on enhancing performance as well as the practicality, interpretability, and user-friendliness. By aligning the specific requirements of various scenarios with the characteristics of associated algorithms, it is possible to devise more accurate, efficient, and comprehensible IQA methods, thereby facilitating the development and application of IQA technology.
\bibliographystyle{named}
\bibliography{ijcai25}

\begin{thebibliography}{}

\bibitem[\protect\citeauthoryear{Aydin \bgroup \em et al.\egroup }{2008}]{aydin2008dynamic}
Tun{\c{c}}~Ozan Aydin, Hans-Peter Seidel, and et~al.
\newblock Dynamic range independent image quality assessment.
\newblock {\em ACM TOG}, 27(3):1--10, 2008.

\bibitem[\protect\citeauthoryear{Bosse and et al.}{2017}]{bosse2017deep}
Sebastian Bosse and et~al.
\newblock Deep neural networks for no-reference and full-reference image quality assessment.
\newblock {\em IEEE TIP}, 27(1):206--219, 2017.

\bibitem[\protect\citeauthoryear{Chahine \bgroup \em et al.\egroup }{2024}]{chahine2024deep}
Nicolas Chahine, others, and et~al.
\newblock Deep portrait quality assessment. a ntire 2024 challenge survey.
\newblock In {\em CVPR}, pages 6732--6744, 2024.

\bibitem[\protect\citeauthoryear{Chandler and Hemami}{}]{chandler2007vsnr}
Damon~M Chandler and Sheila~S Hemami.
\newblock Vsnr: A wavelet-based visual signal-to-noise ratio for natural images.
\newblock {\em IEEE TIP}, (9):2284--2298.

\bibitem[\protect\citeauthoryear{Chang \bgroup \em et al.\egroup }{2013}]{chang2013sparse}
Hua-Wen Chang, Ming-Hui Wang, and et~al.
\newblock Sparse feature fidelity for perceptual image quality assessment.
\newblock {\em IEEE TIP}, 22(10):4007--4018, 2013.

\bibitem[\protect\citeauthoryear{Cheon \bgroup \em et al.\egroup }{2021}]{cheon2021perceptual}
Manri Cheon, Junwoo Lee, and et~al.
\newblock Perceptual image quality assessment with transformers.
\newblock In {\em CVPR}, pages 433--442, 2021.

\bibitem[\protect\citeauthoryear{Ding \bgroup \em et al.\egroup }{2020}]{ding2020image}
Keyan Ding, Eero~P Simoncelli, and et~al.
\newblock Image quality assessment: Unifying structure and texture similarity.
\newblock {\em IEEE TPAMI}, 44(5):2567--2581, 2020.

\bibitem[\protect\citeauthoryear{Egiazarian and et al.}{2006}]{egiazarian2006new}
Karen Egiazarian and et~al.
\newblock New full-reference quality metrics based on hvs.
\newblock In {\em Proceedings of the second international workshop on video processing and quality metrics}, volume~4, page~4, 2006.

\bibitem[\protect\citeauthoryear{Ferzli and Karam}{2009}]{ferzli2009no}
Rony Ferzli and Lina~J Karam.
\newblock A no-reference objective image sharpness metric based on the notion of just noticeable blur (jnb).
\newblock {\em IEEE TIP}, 18(4):717--728, 2009.

\bibitem[\protect\citeauthoryear{Gabarda and Crist{\'o}bal}{2007}]{gabarda2007blind}
Salvador Gabarda and Gabriel Crist{\'o}bal.
\newblock Blind image quality assessment through anisotropy.
\newblock {\em JOSA A}, 24(12):B42--B51, 2007.

\bibitem[\protect\citeauthoryear{Golestaneh \bgroup \em et al.\egroup }{2022}]{golestaneh2022no}
S~Alireza Golestaneh, Kris~M Kitani, and et~al.
\newblock No-reference image quality assessment via transformers, relative ranking, and self-consistency.
\newblock In {\em WACV}, pages 1220--1230, 2022.

\bibitem[\protect\citeauthoryear{Gore and Gupta}{2015}]{gore2015full}
Akshay Gore and Savita Gupta.
\newblock Full reference image quality metrics for jpeg compressed images.
\newblock {\em AEU-International Journal of Electronics and Communications}, 69(2):604--608, 2015.

\bibitem[\protect\citeauthoryear{Gu and et al.}{2022}]{gu2022ntire}
Jinjin Gu and et~al.
\newblock Ntire 2022 challenge on perceptual image quality assessment.
\newblock In {\em CVPR}, pages 951--967, 2022.

\bibitem[\protect\citeauthoryear{Gu \bgroup \em et al.\egroup }{2017}]{gu2017fast}
Ke~Gu, Weisi Lin, and et~al.
\newblock A fast reliable image quality predictor by fusing micro-and macro-structures.
\newblock {\em IEEE Transactions on Industrial Electronics}, 64(5):3903--3912, 2017.

\bibitem[\protect\citeauthoryear{Guo \bgroup \em et al.\egroup }{2021}]{guo2021iqma}
Haiyang Guo, Hengliang Luo, and et~al.
\newblock Iqma network: Image quality multi-scale assessment network.
\newblock In {\em CVPR}, pages 443--452, 2021.

\bibitem[\protect\citeauthoryear{Kang \bgroup \em et al.\egroup }{2014}]{Kang_2014_CVPR}
Le~Kang, David Doermann, and et~al.
\newblock Convolutional neural networks for no-reference image quality assessment.
\newblock In {\em CVPR}, June 2014.

\bibitem[\protect\citeauthoryear{Ke \bgroup \em et al.\egroup }{2021}]{ke2021musiq}
Junjie Ke, Feng Yang, and et~al.
\newblock Musiq: Multi-scale image quality transformer.
\newblock In {\em ICCV}, pages 5148--5157, 2021.

\bibitem[\protect\citeauthoryear{Kim and Lee}{2016}]{kim2016fully}
Jongyoo Kim and Sanghoon Lee.
\newblock Fully deep blind image quality predictor.
\newblock {\em IEEE JSTSP}, 11(1):206--220, 2016.

\bibitem[\protect\citeauthoryear{Kim \bgroup \em et al.\egroup }{2018}]{kim2018deep}
Jongyoo Kim, Sanghoon Lee, and et~al.
\newblock Deep cnn-based blind image quality predictor.
\newblock {\em IEEE TNNLS}, 30(1):11--24, 2018.

\bibitem[\protect\citeauthoryear{Larson and Chandler}{2010}]{larson2010most}
Eric~C Larson and Damon~M Chandler.
\newblock Most apparent distortion: full-reference image quality assessment and the role of strategy.
\newblock {\em JEI}, 2010.

\bibitem[\protect\citeauthoryear{Li \bgroup \em et al.\egroup }{2015}]{li2015no}
Leida Li, Alex~C Kot, and et~al.
\newblock No-reference image blur assessment based on discrete orthogonal moments.
\newblock {\em IEEE TCYB}, 46(1):39--50, 2015.

\bibitem[\protect\citeauthoryear{Li \bgroup \em et al.\egroup }{2016}]{li2016sparse}
Leida Li, Xingming Sun, and et~al.
\newblock Sparse representation-based image quality index with adaptive sub-dictionaries.
\newblock {\em IEEE TIP}, 25(8):3775--3786, 2016.

\bibitem[\protect\citeauthoryear{Lin and Wang}{2018}]{lin2018hallucinated}
Kwan-Yee Lin and Guanxiang Wang.
\newblock Hallucinated-iqa: No-reference image quality assessment via adversarial learning.
\newblock In {\em CVPR}, pages 732--741, 2018.

\bibitem[\protect\citeauthoryear{Lin \bgroup \em et al.\egroup }{2020}]{lin2020deepfl}
Hanhe Lin, Dietmar Saupe, and et~al.
\newblock Deepfl-iqa: Weak supervision for deep iqa feature learning.
\newblock {\em arXiv preprint arXiv:2001.08113}, 2020.

\bibitem[\protect\citeauthoryear{Liu \bgroup \em et al.\egroup }{2011}]{liu2011image}
Anmin Liu, Manish Narwaria, and et~al.
\newblock Image quality assessment based on gradient similarity.
\newblock {\em IEEE TIP}, 21(4):1500--1512, 2011.

\bibitem[\protect\citeauthoryear{Liu \bgroup \em et al.\egroup }{2012}]{liu2012image}
Tsung-Jung Liu, C-C~Jay Kuo, and et~al.
\newblock Image quality assessment using multi-method fusion.
\newblock {\em IEEE TIP}, 22(5):1793--1807, 2012.

\bibitem[\protect\citeauthoryear{Liu \bgroup \em et al.\egroup }{2013}]{liu2013no}
Yiming Liu, Szymon Rusinkiewicz, and et~al.
\newblock A no-reference metric for evaluating the quality of motion deblurring.
\newblock {\em ACM TOG}, 2013.

\bibitem[\protect\citeauthoryear{Liu \bgroup \em et al.\egroup }{2015}]{liu2015paraboost}
Tsung-Jung Liu, C-C~Jay Kuo, and et~al.
\newblock A paraboost method to image quality assessment.
\newblock {\em IEEE TNNLS}, 28(1):107--121, 2015.

\bibitem[\protect\citeauthoryear{Liu \bgroup \em et al.\egroup }{2017}]{liu2017rankiqa}
Xialei Liu, Andrew~D Bagdanov, and et~al.
\newblock Rankiqa: Learning from rankings for no-reference image quality assessment.
\newblock In {\em ICCV}, pages 1040--1049, 2017.

\bibitem[\protect\citeauthoryear{Luo and Tang}{2008}]{luo2008photo}
Yiwen Luo and Xiaoou Tang.
\newblock Photo and video quality evaluation: Focusing on the subject.
\newblock In {\em ECCV}, pages 386--399. Springer, 2008.

\bibitem[\protect\citeauthoryear{Ma \bgroup \em et al.\egroup }{2017}]{ma2017end}
Kede Ma, Wangmeng Zuo, and et~al.
\newblock End-to-end blind image quality assessment using deep neural networks.
\newblock {\em IEEE TIP}, 27(3):1202--1213, 2017.

\bibitem[\protect\citeauthoryear{Madhusudana \bgroup \em et al.\egroup }{2022}]{madhusudana2022image}
Pavan~C Madhusudana, Alan~C Bovik, and et~al.
\newblock Image quality assessment using contrastive learning.
\newblock {\em IEEE TIP}, 31:4149--4161, 2022.

\bibitem[\protect\citeauthoryear{Min \bgroup \em et al.\egroup }{2019}]{min2019quality}
Xiongkuo Min, Wenjun Zhang, and et~al.
\newblock Quality evaluation of image dehazing methods using synthetic hazy images.
\newblock {\em IEEE TMM}, 21(9):2319--2333, 2019.

\bibitem[\protect\citeauthoryear{Mittal \bgroup \em et al.\egroup }{2012}]{mittal2012no}
Anish Mittal, Alan~Conrad Bovik, and et~al.
\newblock No-reference image quality assessment in the spatial domain.
\newblock {\em IEEE TIP}, 21(12):4695--4708, 2012.

\bibitem[\protect\citeauthoryear{Moorthy and Bovik}{2011}]{moorthy2011blind}
Anush~Krishna Moorthy and Alan~Conrad Bovik.
\newblock Blind image quality assessment: From natural scene statistics to perceptual quality.
\newblock {\em IEEE TIP}, 20(12):3350--3364, 2011.

\bibitem[\protect\citeauthoryear{Rosen \bgroup \em et al.\egroup }{2018}]{rosen2018quantitative}
Adon~FG Rosen, others, and et~al.
\newblock Quantitative assessment of structural image quality.
\newblock {\em Neuroimage}, 169:407--418, 2018.

\bibitem[\protect\citeauthoryear{Royat \bgroup \em et al.\egroup }{2021}]{royat2021ntire}
Ali Royat, others, and et~al.
\newblock Ntire 2021 challenge on perceptual image quality assessment.
\newblock In {\em CVPR}. IEEE, 2021.

\bibitem[\protect\citeauthoryear{Saad \bgroup \em et al.\egroup }{2012}]{saad2012blind}
Michele~A Saad, Christophe Charrier, and et~al.
\newblock Blind image quality assessment: A natural scene statistics approach in the dct domain.
\newblock {\em IEEE TIP}, 21(8):3339--3352, 2012.

\bibitem[\protect\citeauthoryear{Saha \bgroup \em et al.\egroup }{2023}]{saha2023re}
Avinab Saha, Alan~C Bovik, and et~al.
\newblock Re-iqa: Unsupervised learning for image quality assessment in the wild.
\newblock In {\em CVPR}, pages 5846--5855, 2023.

\bibitem[\protect\citeauthoryear{Series}{2012}]{series2012methodology}
B~Series.
\newblock Methodology for the subjective assessment of the quality of television pictures.
\newblock {\em ITU-R BT}, 500(13), 2012.

\bibitem[\protect\citeauthoryear{Sheikh and Bovik}{2006}]{sheikh2006image}
Hamid~R Sheikh and Alan~C Bovik.
\newblock Image information and visual quality.
\newblock {\em IEEE TIP}, 15(2):430--444, 2006.

\bibitem[\protect\citeauthoryear{Sheikh \bgroup \em et al.\egroup }{2005}]{sheikh2005information}
Hamid~R Sheikh, Gustavo De~Veciana, and et~al.
\newblock An information fidelity criterion for image quality assessment using natural scene statistics.
\newblock {\em IEEE TIP}, 14(12):2117--2128, 2005.

\bibitem[\protect\citeauthoryear{Shnayderman \bgroup \em et al.\egroup }{2006}]{shnayderman2006svd}
Aleksandr Shnayderman, Ahmet~M Eskicioglu, and et~al.
\newblock An svd-based grayscale image quality measure for local and global assessment.
\newblock {\em IEEE TIP}, 15(2):422--429, 2006.

\bibitem[\protect\citeauthoryear{Su \bgroup \em et al.\egroup }{2020}]{su2020blindly}
Shaolin Su, Yanning Zhang, and et~al.
\newblock Blindly assess image quality in the wild guided by a self-adaptive hyper network.
\newblock In {\em CVPR}, 2020.

\bibitem[\protect\citeauthoryear{Thung and Raveendran}{2009}]{5412098}
Kim-Han Thung and Paramesran Raveendran.
\newblock A survey of image quality measures.
\newblock In {\em 2009 International Conference for Technical Postgraduates (TECHPOS)}, pages 1--4, 2009.

\bibitem[\protect\citeauthoryear{Varga}{2020}]{varga2020multi}
Domonkos Varga.
\newblock Multi-pooled inception features for no-reference image quality assessment.
\newblock {\em Appl. Sci.}, 10(6):2186, 2020.

\bibitem[\protect\citeauthoryear{Wang and Li}{2010}]{wang2010information}
Zhou Wang and Qiang Li.
\newblock Information content weighting for perceptual image quality assessment.
\newblock {\em IEEE TIP}, 20(5):1185--1198, 2010.

\bibitem[\protect\citeauthoryear{Wang and Simoncelli}{2005}]{wang2005reduced}
Zhou Wang and Eero~P Simoncelli.
\newblock Reduced-reference image quality assessment using a wavelet-domain natural image statistic model.
\newblock In {\em HVEI}, volume 5666, pages 149--159. SPIE, 2005.

\bibitem[\protect\citeauthoryear{Wang \bgroup \em et al.\egroup }{2002}]{wang2002image}
Zhou Wang, Ligang Lu, and et~al.
\newblock Why is image quality assessment so difficult?
\newblock In {\em ICASSP}, volume~4, pages IV--3313. IEEE, 2002.

\bibitem[\protect\citeauthoryear{Wang \bgroup \em et al.\egroup }{2003}]{wang2003multiscale}
Zhou Wang, Alan~C Bovik, and et~al.
\newblock Multiscale structural similarity for image quality assessment.
\newblock In {\em ACSSC}, volume~2, pages 1398--1402. Ieee, 2003.

\bibitem[\protect\citeauthoryear{Wang \bgroup \em et al.\egroup }{2004}]{wang2004image}
Zhou Wang, Eero~P Simoncelli, and et~al.
\newblock Image quality assessment: from error visibility to structural similarity.
\newblock {\em IEEE TIP}, 13(4):600--612, 2004.

\bibitem[\protect\citeauthoryear{Wang}{2002}]{wang2002universal}
Zhou Wang.
\newblock A universal image quality index.
\newblock {\em IEEE SPL}, 9(3):81--84, 2002.

\bibitem[\protect\citeauthoryear{Xu \bgroup \em et al.\egroup }{2017}]{Xu2017NoreferenceBlindIQ}
Shaoping Xu, Weidong Min, and et~al.
\newblock No-reference/blind image quality assessment: A survey.
\newblock {\em IETE Technical Review}, 34:223 -- 245, 2017.

\bibitem[\protect\citeauthoryear{Xue \bgroup \em et al.\egroup }{2013a}]{xue2013gradient}
Wufeng Xue, Alan~C Bovik, and et~al.
\newblock Gradient magnitude similarity deviation: A highly efficient perceptual image quality index.
\newblock {\em IEEE TIP}, 23(2):684--695, 2013.

\bibitem[\protect\citeauthoryear{Xue \bgroup \em et al.\egroup }{2013b}]{xue2013learning}
Wufeng Xue, Xuanqin Mou, and et~al.
\newblock Learning without human scores for blind image quality assessment.
\newblock In {\em CVPR}, pages 995--1002, 2013.

\bibitem[\protect\citeauthoryear{Yang \bgroup \em et al.\egroup }{2019}]{8822415}
Xiaohan Yang, Hantao Liu, and et~al.
\newblock A survey of dnn methods for blind image quality assessment.
\newblock {\em IEEE Access}, 7:123788--123806, 2019.

\bibitem[\protect\citeauthoryear{Yang \bgroup \em et al.\egroup }{2022}]{yang2022maniqa}
Sidi Yang, Yujiu Yang, and et~al.
\newblock Maniqa: Multi-dimension attention network for no-reference image quality assessment.
\newblock In {\em CVPR}, pages 1191--1200, 2022.

\bibitem[\protect\citeauthoryear{Yeganeh and Wang}{2012}]{yeganeh2012objective}
Hojatollah Yeganeh and Zhou Wang.
\newblock Objective quality assessment of tone-mapped images.
\newblock {\em IEEE TIP}, 22(2):657--667, 2012.

\bibitem[\protect\citeauthoryear{Yin \bgroup \em et al.\egroup }{2022}]{yin2022content}
Guanghao Yin, Changhu Wang, and et~al.
\newblock Content-variant reference image quality assessment via knowledge distillation.
\newblock In {\em AAAI}, volume~36, pages 3134--3142, 2022.

\bibitem[\protect\citeauthoryear{You and Korhonen}{2021}]{you2021transformer}
Junyong You and Jari Korhonen.
\newblock Transformer for image quality assessment.
\newblock In {\em ICIP}, pages 1389--1393. IEEE, 2021.

\bibitem[\protect\citeauthoryear{Zhang \bgroup \em et al.\egroup }{2011}]{zhang2011fsim}
Lin Zhang, David Zhang, and et~al.
\newblock Fsim: A feature similarity index for image quality assessment.
\newblock {\em IEEE TIP}, 20(8):2378--2386, 2011.

\bibitem[\protect\citeauthoryear{Zhang \bgroup \em et al.\egroup }{2014}]{zhang2014vsi}
Lin Zhang, Hongyu Li, and et~al.
\newblock Vsi: A visual saliency-induced index for perceptual image quality assessment.
\newblock {\em IEEE TIP}, 23(10):4270--4281, 2014.

\bibitem[\protect\citeauthoryear{Zhang \bgroup \em et al.\egroup }{2015}]{zhang2015feature}
Lin Zhang, Alan~C Bovik, and et~al.
\newblock A feature-enriched completely blind image quality evaluator.
\newblock {\em IEEE TIP}, 24(8):2579--2591, 2015.

\bibitem[\protect\citeauthoryear{Zhang \bgroup \em et al.\egroup }{2018}]{zhang2018unreasonable}
Richard Zhang, Oliver Wang, and et~al.
\newblock The unreasonable effectiveness of deep features as a perceptual metric.
\newblock In {\em CVPR}, pages 586--595, 2018.

\bibitem[\protect\citeauthoryear{Zhang \bgroup \em et al.\egroup }{2021a}]{zhang2021cnn}
Siyuan Zhang, Wenguang Hou, and et~al.
\newblock Cnn-based medical ultrasound image quality assessment.
\newblock {\em Complexity}, 2021(1):9938367, 2021.

\bibitem[\protect\citeauthoryear{Zhang \bgroup \em et al.\egroup }{2021b}]{zhang2021uncertainty}
Weixia Zhang, Xiaokang Yang, and et~al.
\newblock Uncertainty-aware blind image quality assessment in the laboratory and wild.
\newblock {\em IEEE TIP}, 30:3474--3486, 2021.

\bibitem[\protect\citeauthoryear{Zhu \bgroup \em et al.\egroup }{2020}]{zhu2020metaiqa}
Hancheng Zhu, Guangming Shi, and et~al.
\newblock Metaiqa: Deep meta-learning for no-reference image quality assessment.
\newblock In {\em CVPR}, pages 14143--14152, 2020.

\end{thebibliography}
\end{document}